\documentclass[sigconf,nonacm]{acmart}
\setcopyright{none}
\acmConference{}{}{}
\acmDOI{}
\acmISBN{}
\acmYear{2026}
\copyrightyear{2026}
\usepackage{amsmath}
\usepackage{colortbl}
\definecolor{WeightLoopTint}{HTML}{EAF3FA}
\newcommand{\BU}{\mathrm{BU}}

\begin{document}
\title{WeightLoop: Weight-Space Programming with Experimental Feedback}
\author{Mengran Li}
\author{Bo Li}
\author{Weijing Zhao}
\author{Chengyang Zhang}
\author{Boyu Zhang}
\author{Jinfeng Xu}
\author{Zelin Zang}
\author{Victor C. M. Leung}
\author{Stan Z. Li}
\authorsaddresses{}
\renewcommand{\shortauthors}{Mengran Li et al.}
\makeatletter
\def\@mkauthors{%
  \global\setbox\mktitle@bx=\vbox{\unvbox\mktitle@bx
  \centering
  \begin{minipage}{\textwidth}
  \centering
  {\fontsize{11}{14}\selectfont
  Mengran Li\textsuperscript{1},\quad Bo Li\textsuperscript{2},\quad
  Weijing Zhao\textsuperscript{2},\quad Chengyang Zhang\textsuperscript{3},\quad
  Boyu Zhang\textsuperscript{4}\\[2pt]
  Jinfeng Xu\textsuperscript{5,*},\quad Zelin Zang\textsuperscript{1,*},\quad
  Victor C. M. Leung\textsuperscript{5},\quad Stan Z. Li\textsuperscript{6}\par}
  \vspace{4pt}
  {\fontsize{9}{11.5}\selectfont
  \textsuperscript{1}Hong Kong Institute of Science and Innovation\quad
  \textsuperscript{2}University of Macau\\
  \textsuperscript{3}Sichuan University\quad
  \textsuperscript{4}Sun Yat-sen University\\
  \textsuperscript{5}The University of British Columbia\quad
  \textsuperscript{6}Westlake University\par}
  \vspace{3pt}
  {\fontsize{9}{11}\selectfont\href{mailto:limengran1998@163.com}{limengran1998@163.com}\par}
  \vspace{2pt}
  {\fontsize{8.5}{10.5}\selectfont\textsuperscript{*}Co-corresponding authors: Jinfeng Xu and Zelin Zang.\par}
  \end{minipage}\par
  \vspace{10pt}}}
\makeatother
\hypersetup{pdfauthor={Mengran Li; Bo Li; Weijing Zhao; Chengyang Zhang; Boyu Zhang; Jinfeng Xu; Zelin Zang; Victor C. M. Leung; Stan Z. Li},
 pdftitle={WeightLoop: Weight-Space Programming with Experimental Feedback},
 pdfsubject={Public author manuscript},pdfkeywords={WeightLoop, model editing, experimental feedback, weight-space programming}}

\begin{abstract}
Scientific predictors guide costly experiments, yet returned measurements often reveal errors that matter for subsequent experimental choices. WeightLoop turns this feedback into persistent model updates through a reusable Weight-Space Programming (WSP) editor. We use GPT-6 Astra and DeepSeek to construct update programs from the scientific objective, model computation and editable parameters. These programs use accumulated measurements to estimate parameter corrections, validate them on feedback and save accepted changes into the original model. With identical feedback and writable parameters, four pretrained scientific predictors obtain 5.10--18.56\% lower primary error. For weather prediction, changing four output-bias parameters reduces root mean squared error by 13.28\%. Functional ranking and sequential discovery further connect reusable editing programs to improved experimental choices. Code: \url{https://github.com/limengran98/WeightLoop}.

\end{abstract}

\maketitle

\section{Introduction}
Scientific predictors determine which costly experiments are performed and which evidence becomes available next~\cite{activelearning,robotscientist,biodiscovery}. A deployed predictor initially has few relevant assays. Its selected experiments return observations, including discrepancies from predictions, that should improve the model used for the next choice. The challenge is to turn a small, growing set of measurements into useful changes the model's parameters can express.

Useful updates must respect the scientific objective and the deployed computation. Model editing changes network behavior~\cite{editable,mend,knowledgeeditor} and factual associations~\cite{rome,memit}. Weight-Space Programming (WSP) shows that language models can inspect and modify weights~\cite{takehi2026weightspace}; WeightCode installs computations from task knowledge through reusable programs~\cite{weightcode2026}. Scientific feedback makes updating recurrent: predictions select measurements, and those measurements inform the next predictions and choices.

WeightLoop uses GPT-6 Astra~\cite{openai2026astra} to construct a program that translates measured discrepancies into native parameter corrections. GPT-6 Astra inspects the objective, forward computation, editable parameters and validation interface; returned measurements determine the numerical correction. We also construct programs independently with DeepSeek~\cite{deepseek2026api}. Both workflows supply update code to the same feedback-validation and deployment interface.

\begin{figure*}[t]
\centering
\includegraphics[width=\linewidth]{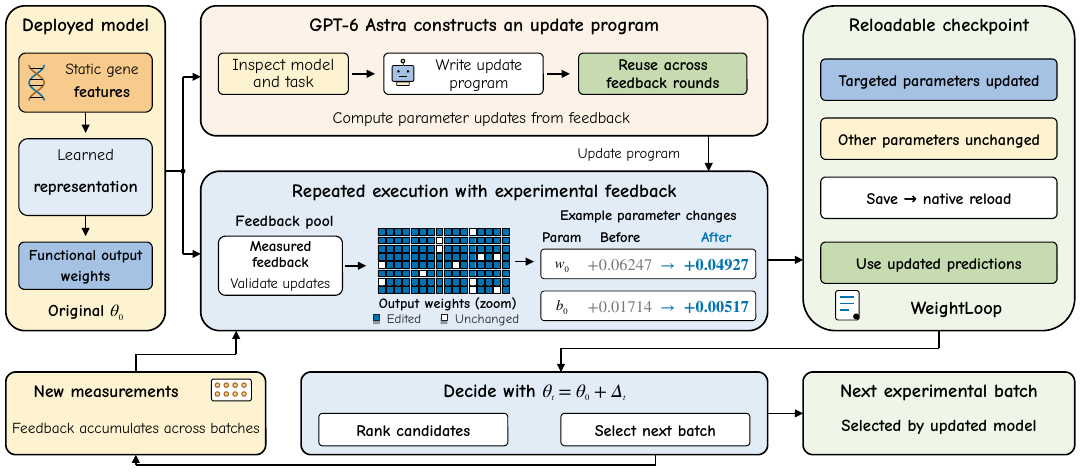}
\caption{Feedback updates the predictor used to choose the next experiments. The HepG2 grid zooms into 128 output weights: 116 blue cells are edited and 12 white cells unchanged. Including 128 hidden biases, the 244 edited parameters account for 0.0068\% of the model. $w_0$ and $b_0$ denote the first output weight and hidden bias.}
\Description{GPT-6 Astra constructs a reusable editor. Returned measurements determine native parameter changes, and the updated model guides the next experimental batch.}
\label{fig:workflow}
\end{figure*}

Figure~\ref{fig:workflow} shows feedback returning to the predictor. Each round estimates a total correction from the original model and accumulated measurements. The updated predictor chooses the next batch, whose results extend the feedback used by the program.

To test whether feedback improves scientific predictions, we compare updates under identical feedback and writable parameters in cell biology~\cite{gears}, earthquake monitoring~\cite{phasenet}, weather prediction~\cite{climax} and fluid dynamics~\cite{fno}. Functional ranking and sequential discovery then evaluate which experiments the updated models select. RNA calibration tests the same approach on larger decoders. Together, these studies examine how experimental measurements improve predictions and subsequent decisions.

\section{WeightLoop}
\subsection{From a scientific objective to an update program}
New measurements help when the deployed predictor can express what they reveal. We provide GPT-6 Astra with the task, model code and permitted parameters in an interactive coding session to build an update program. Separately, we send these inputs to DeepSeek through its official API to generate another program. We check each program's input and output formats on synthetic data. Subsequent rounds run the program on returned measurements to fit, validate and save parameter updates.

Let $f_\theta$ be the native predictor with parameter vector $\theta$, and let $\theta_0$ denote its original parameters. At feedback round $t=1,2,\ldots$, the accumulated measurements are $S_t=\{(x_i,y_i)\}_{i=1}^{n_t}$, where $x_i$ is an experimental input, $y_i$ its measured outcome, and $n_t$ the number of returned observations. A reusable program $P$ proposes
\begin{equation}
 \Delta=P(\theta_0,S_t;\eta),\qquad
 \Delta_j=0\quad(j\notin\mathcal I),
\end{equation}
where $\Delta$ is the correction vector, $j$ indexes a parameter coordinate, $\mathcal I$ is the permitted coordinate set, and $\eta\in\mathcal H$ selects a numerical configuration from a fixed menu $\mathcal H$.

The program describes how permitted parameter changes affect predictions. Each round supplies new measurements to this computation, which estimates a correction and evaluates it through the model.

\subsection{Selecting and committing a correction}
A correction must help beyond the observations used to fit it. WeightLoop divides returned measurements into $V$ feedback folds, with fitting subset $S_{t,k}^{\rm fit}$ and validation subset $S_{t,k}^{\rm val}$ for fold $k\in\{1,\ldots,V\}$. The task loss $\mathcal L(f,S)$ evaluates predictor $f$ on measurements $S$. Let $\mathcal H_t\subseteq\mathcal H$ contain settings passing the fixed acceptance checks. When this set is nonempty, WeightLoop selects
\begin{equation}
 \eta_t^*=\arg\min_{\eta\in\mathcal H_t}
 \sum_{k=1}^{V}\mathcal L\!\left(
 f_{\theta_0+P(\theta_0,S_{t,k}^{\rm fit};\eta)},
 S_{t,k}^{\rm val}\right),
\end{equation}
where $\eta_t^*$ is the selected configuration. This evaluates proposals through the native forward computation.

Using the selected configuration, the program refits a correction on all accumulated feedback and writes it into the model:
\begin{equation}
 \theta_t=\theta_0+P(\theta_0,S_t;\eta_t^*),\qquad
 A_{t+1}=\mathcal D(f_{\theta_t}).
\end{equation}
Here $\theta_t$ is the deployed parameter vector, $\mathcal D$ is the experimental-selection rule applied to the remaining candidates, and $A_{t+1}$ is the next experimental batch. Its returned outcomes extend $S_t$. If no setting passes acceptance, WeightLoop retains the previous deployment $\theta_{t-1}$. Every total correction is reconstructed against $\theta_0$, so accumulating evidence revises one common reference. Native save and reload reproduce the accepted predictor before it guides another batch.

\section{Experiments}
\subsection{Tasks and common evaluation settings}
\paragraph{Four prediction tasks.}
We evaluate four prediction tasks (Table~\ref{tab:static}). \emph{Cell responses:} GEARS~\cite{gears} predicts gene-expression changes after gene silencing in Adamson CRISPRi~\cite{adamson}. We measure mean squared error (MSE) on nine target genes and 1,024 RNA outputs selected from untreated controls. \emph{Earthquake timing:} PhaseNet~\cite{phasenet} predicts when a seismic P wave arrives in 64 whole INSTANCE~\cite{instance} events. Mean absolute error (MAE, seconds) uses the 100\,Hz grid and includes events without a thresholded pick. \emph{Weather prediction:} ClimaX~\cite{climax} predicts the 500\,hPa geopotential field, which describes upper-air circulation. We measure area-weighted root mean squared error (RMSE, $\mathrm{m^2\,s^{-2}}$) on 59 ERA5 Z500 states in six whole months~\cite{weatherbench2}. \emph{Shallow-water flow:} a two-dimensional Fourier neural operator (FNO2d)~\cite{fno} predicts water-height fields after a radial dam break in PDEBench~\cite{pdebench}. Relative $L_2$ error compares predicted and observed fields over 32 initial-state groups, using ten observed fields; $L_2$ is the Euclidean field norm. The published split includes training-equivalent initial states.

The four tasks use execution seed17 and a regularized numerical editor within DeepSeek-selected parameter pools. WeightLoop and Adam~\cite{adam2015} start from the same model and receive the same ordered feedback, writable coordinates, validation folds and acceptance rules. We add feedback in three batches: cumulative totals are 8/12/16 genes, 24/40/56 earthquakes, 20/41/61 weather states and 10/18/26 simulation groups. The respective editable pools contain 1,024/27/192/129 parameters. Each update is fitted before the next batch is revealed. Final evaluation objects remain outside feedback fitting.

\begin{table*}[t]
\centering\small\setlength{\tabcolsep}{5pt}
\caption{Prediction with identical feedback and writable parameters. Each error metric is lower-is-better ($\downarrow$). $E_{\rm Adam}$ and $E_{\rm WL}$ are the reported task errors; relative improvement is $100(E_{\rm Adam}-E_{\rm WL})/E_{\rm Adam}$.}
\label{tab:static}
\begin{tabular}{@{}lllrrr@{}}
\toprule
Task&Predictor&Metric&Adam&WeightLoop&Improvement (\%)\\
\midrule
Cell responses & GEARS & RNA MSE $\downarrow$ & $4.496\!\times\!10^{-2}$ & \cellcolor{WeightLoopTint}$\mathbf{4.258\!\times\!10^{-2}}$ & 5.31 \\
Earthquake timing & PhaseNet & P MAE (s) $\downarrow$ & $6.446$ & \cellcolor{WeightLoopTint}$\mathbf{5.250}$ & 18.56 \\
Weather prediction & ClimaX & RMSE ($\mathrm{m^2\,s^{-2}}$) $\downarrow$ & $1187.58$ & \cellcolor{WeightLoopTint}$\mathbf{1029.83}$ & 13.28 \\
Shallow-water flow & FNO2d & Relative $L_2\downarrow$ & $8.467\!\times\!10^{-3}$ & \cellcolor{WeightLoopTint}$\mathbf{8.035\!\times\!10^{-3}}$ & 5.10 \\
\bottomrule
\end{tabular}
\end{table*}

\paragraph{Choosing new experiments.}
We also test which genes a model recommends for new experiments. Four biological tasks seek reduced growth in HepG2 and Jurkat cells using negative Chronos effects from DepMap~\cite{depmap24q4,chronos}, or reduced HL60 proliferation and migration~\cite{belliveau}. Three sequential tasks seek higher LC3/NBR1 autophagic flux, a measure of cellular recycling~\cite{shoemaker}, or lower HuH7 uptake of LDL cholesterol~\cite{emmer}. Each predictor is a multilayer perceptron (MLP) using GO~\cite{geneontology}, scGPT~\cite{scgpt} and MAP-KG~\cite{mapkg} gene features. Editing changes its final hidden biases and output weights. The initial MLP has 25 training labels. Single-update editing receives 75 additional assays; sequential policies select five batches of 20 genes and use their own returned results. Single-update evaluation genes stay outside feedback fitting. We average five gene folds within each of three model-training seeds, then report the mean and sample standard deviation across those three seed means.

\paragraph{RNA calibration.}
An H1 CRISPRi study uses scGPT~\cite{scgpt} and scFoundation with GEARS (scF+GEARS)~\cite{scfoundation,gears}. Three fixed partitions per training seed contain 40 adaptation, 16 feedback and 20 query targets. Adam and the GPT-6 Astra editor share the adapted checkpoint, the 16 feedback targets, writable tensors and two feedback-validation folds. We report overall MSE across three training seeds after averaging their three partitions.

\paragraph{Choosing and fitting an update.}
For the four prediction tasks, we test each update on feedback withheld from fitting. We accept it when validation loss improves by at least 0.1\% over the initial model and loss on observed history worsens by no more than 5\% relative to the current model. Otherwise, we keep the current model. Accepted updates are saved in the model's float32 format before predicting the next batch.

We test 12 preset fitting settings. They combine squared error or smooth MAE with two penalties: one keeps predictions on reference inputs close to the original model; the other limits the parameter change. Squared-error settings use regularized least squares~\cite{ridge}. Each feedback gene, event, weather state or simulation group receives equal weight, averaged over its outputs. Output and parameter scales make the fit comparable across quantities with different units.

For linear parameter-to-output relationships, the editor computes the effect of a change exactly. For nonlinear PhaseNet, it approximates this effect to propose an update, then validates the updated model's actual predictions. Smooth MAE is fitted with L-BFGS using $\sqrt{e^2+0.01^2}-0.01$, where $e$ is the normalized prediction error. Adam fits the same 12 objectives from zero correction, with learning rate 0.03, moment coefficients 0.9/0.999 and denominator stabilizer $\epsilon=10^{-8}$. Its best-objective iterate supplies the proposal. Both approaches use the same validation procedure. Bold blue cells mark the best value in each comparison row.

\subsection{Same-feedback updates improve prediction}
\begin{figure*}[t]
\centering
\includegraphics[width=\linewidth]{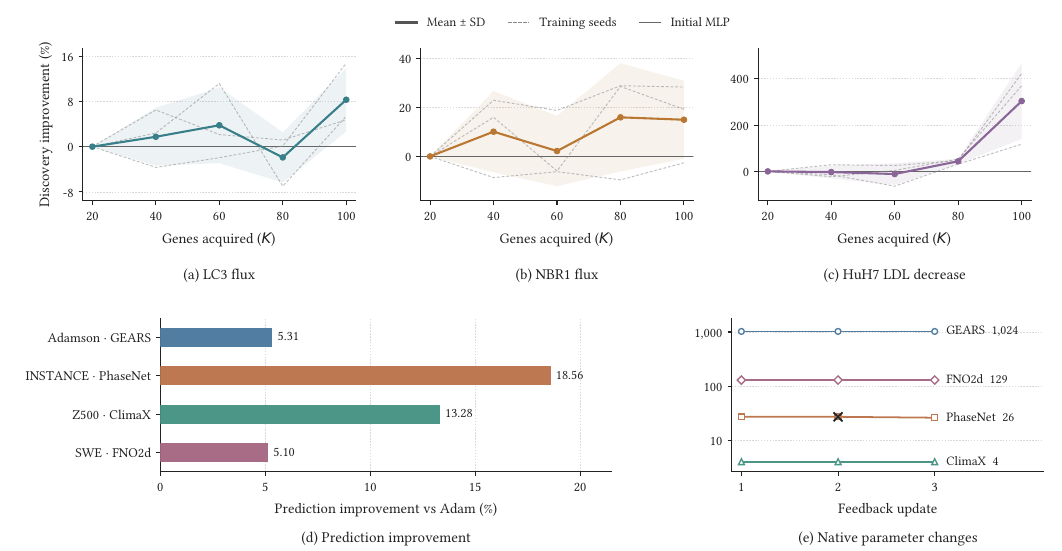}
\caption{Experimental choices and parameter updates. (a--c) Relative discovery improvements from GPT-6 Astra editing. At each budget, paired seed differences are expressed as a percentage of the initial MLP policy's three-seed mean utility. Dashed lines show the three training seeds; solid lines and bands show their mean $\pm$ sample SD, after averaging five gene folds per seed. (d) Relative prediction improvement over Adam with identical feedback and writable parameters, using DeepSeek-selected pools. (e) Parameters changed from the initial model (log scale); crosses mark retained models. Colors match (d).}
\Description{Three sequential experimental-selection trajectories, error reductions for four pretrained scientific predictors, and native parameter changes across feedback updates.}
\label{fig:c}
\end{figure*}

Table~\ref{tab:static} shows lower primary error in every matched-feedback task: relative improvements range from 5.10\% for FNO2d to 18.56\% for PhaseNet. Figure~\ref{fig:c}(d) expresses these task-specific improvements on a common percentage scale.

Repeated updates change native parameters (Figure~\ref{fig:c}(e)): final counts are 1,024 for GEARS, 26 for PhaseNet, four for ClimaX and 129 for FNO2d. PhaseNet retains its deployment on the second prefix and accepts a correction on the third. Native checkpoint reload reproduces the accepted predictions.

\subsection{Feedback improves experimental ranking}
For recommendation budget $B$, let $r\in\{1,\ldots,B\}$ be a rank, $\pi_r$ the candidate at that rank, and $y_{\pi_r}$ its measured outcome, oriented so that larger is preferred. Discount weights satisfy $d_r\propto1/\log_2(r+1)$ and $\sum_{r=1}^{B}d_r=1$. We report
\begin{equation}
 U_B=\sum_{r=1}^{B}d_r y_{\pi_r},\qquad
 \BU_B=\frac{U_B-\bar y_C}{U_B^{\rm oracle}-\bar y_C},
\end{equation}
where $U_B$ is discounted utility, $C$ is the evaluation candidate pool, $\bar y_C$ its mean outcome, and $U_B^{\rm oracle}$ the utility of the best ordering under the measured outcomes. Higher $\BU_B$ is better; zero is same-pool random selection. Score ties use expected utility over tied orders.

Table~\ref{tab:a} shows higher mean utility at both recommendation budgets for every task. The GPT-6 Astra editor uses returned assays to improve the MLP responsible for subsequent recommendations.

\begin{table}[t]
\centering\small\setlength{\tabcolsep}{4pt}
\caption{Functional recommendation, $\BU\uparrow$. All predictors are MLPs. WL-A denotes the editor constructed with GPT-6 Astra. Mean $\pm$ sample SD across three training seeds after averaging five gene folds per seed.}
\label{tab:a}
\begin{tabular}{@{}lrrr@{}}
\toprule
Task&$B$&Initial MLP&WL-A\\
\midrule
HepG2 & 20 & $.5157\!\pm\!.0162$ & \cellcolor{WeightLoopTint}$\boldsymbol{.5785\!\pm\!.0195}$ \\
HepG2 & 50 & $.5069\!\pm\!.0243$ & \cellcolor{WeightLoopTint}$\boldsymbol{.5688\!\pm\!.0177}$ \\
Jurkat & 20 & $.5141\!\pm\!.0217$ & \cellcolor{WeightLoopTint}$\boldsymbol{.5872\!\pm\!.0211}$ \\
Jurkat & 50 & $.5014\!\pm\!.0293$ & \cellcolor{WeightLoopTint}$\boldsymbol{.5834\!\pm\!.0105}$ \\
HL60 prolif. & 20 & $.2132\!\pm\!.0209$ & \cellcolor{WeightLoopTint}$\boldsymbol{.2446\!\pm\!.0026}$ \\
HL60 prolif. & 50 & $.2490\!\pm\!.0113$ & \cellcolor{WeightLoopTint}$\boldsymbol{.2951\!\pm\!.0052}$ \\
HL60 migr. & 20 & $-.0121\!\pm\!.0116$ & \cellcolor{WeightLoopTint}$\boldsymbol{.0361\!\pm\!.0351}$ \\
HL60 migr. & 50 & $.0070\!\pm\!.0059$ & \cellcolor{WeightLoopTint}$\boldsymbol{.0225\!\pm\!.0287}$ \\
\bottomrule
\end{tabular}
\end{table}

\subsection{Repeated updates improve sequential discovery}
Let $K$ be the number of acquired targets, $A_K\subseteq C$ the acquired set, and $g$ a target index with measured outcome $y_g$. Cumulative discovery utility is
\begin{equation}
 D_K=\frac{\sum_{g\in A_K}y_g-K\bar y_C}
 {\sum_{g\in\mathrm{Top}_K(C)}y_g-K\bar y_C},
\end{equation}
where $\mathrm{Top}_K(C)$ contains the $K$ highest-outcome targets in pool $C$. Higher $D_K$ is better; zero is same-pool random acquisition. Each batch is selected before its outcomes return, and each policy follows its own feedback trajectory.

Both GPT-6 Astra and DeepSeek programs improve final mean $D_{100}$ on all three tasks (Table~\ref{tab:c}). Relative to the initial MLP, GPT-6 Astra editing improves mean utility by 8.30\%, 14.97\% and 303.43\% for LC3, NBR1 and HuH7. Relative improvement is $100(\bar D_K^{\rm WL}-\bar D_K^{\rm init})/\bar D_K^{\rm init}$, where the bars denote three-seed means and superscripts identify the edited and initial policies. Figure~\ref{fig:c}(a--c) tracks this improvement as five batches of experimental results arrive.

\begin{table}[t]
\centering\small\setlength{\tabcolsep}{3.5pt}
\caption{Sequential discovery, $D_{100}\uparrow$. All MLP policies acquire 100 genes. Mean $\pm$ sample SD across three training seeds.}
\label{tab:c}
\begin{tabular}{@{}lrrr@{}}
\toprule
Task&Initial MLP&GPT-6 Astra&DeepSeek\\
\midrule
LC3 & $.1910\!\pm\!.0245$ & $.2068\!\pm\!.0213$ & \cellcolor{WeightLoopTint}$\boldsymbol{.2180\!\pm\!.0228}$ \\
NBR1 & $.1621\!\pm\!.0043$ & \cellcolor{WeightLoopTint}$\boldsymbol{.1864\!\pm\!.0268}$ & $.1783\!\pm\!.0127$ \\
HuH7 & $.0097\!\pm\!.0180$ & $.0391\!\pm\!.0025$ & \cellcolor{WeightLoopTint}$\boldsymbol{.0675\!\pm\!.0013}$ \\
\bottomrule
\end{tabular}
\end{table}

\subsection{Feedback calibrates native RNA outputs}
The program estimates a response-amplitude correction and writes it through the native decoder. The correction adjusts the size of the predicted gene-expression response using the feedback targets. Once stored in the decoder, it is applied automatically when the model predicts responses for new target genes. Table~\ref{tab:rna} shows mean MSE reductions of 1.85\% for scGPT and 7.49\% for scF+GEARS relative to their adapted checkpoints. Both update strategies reach close MSE values on scGPT; WeightLoop gives the lowest MSE on scF+GEARS. This supports feedback-driven output calibration on the two RNA architectures.

\begin{table}[tb]
\centering\small\setlength{\tabcolsep}{3pt}
\caption{RNA output calibration, MSE $\downarrow$ in units of $10^{-3}$. Base denotes the adapted checkpoint; WL-A denotes the GPT-6 Astra editor. Mean $\pm$ sample SD across three training-seed means.}
\label{tab:rna}
\begin{tabular}{@{}lrrr@{}}
\toprule
Predictor&Base&Adam&WL-A\\
\midrule
scGPT & $11.204\!\pm\!.182$ & \cellcolor{WeightLoopTint}$\boldsymbol{10.992\!\pm\!.015}$ & $10.997\!\pm\!.013$  \\
scF+GEARS & $12.075\!\pm\!.035$ & $12.075\!\pm\!.035$ & \cellcolor{WeightLoopTint}$\boldsymbol{11.171\!\pm\!.026}$  \\
\bottomrule
\end{tabular}
\end{table}

\section{Conclusion}
WeightLoop uses experimental measurements to improve the model that selects subsequent experiments. Programs constructed with GPT-6 Astra and DeepSeek estimate and validate parameter updates from returned feedback. Functional ranking and sequential discovery improve, and four scientific predictors obtain 5.10--18.56\% lower primary error than Adam with identical feedback. RNA calibration also lowers overall MSE relative to adapted checkpoints. Accepted updates are saved in the original model and reused for subsequent experimental choices.

\section*{Ethical Considerations}
This study reuses public assay, seismic, weather and simulation data. It involves no new studies with human participants or collection of personal information. Recommendations guide research assays. Original checkpoints are retained so updates remain reversible.

\bibliographystyle{ACM-Reference-Format}
\bibliography{references}
\end{document}